\documentclass[runningheads]{llncs}
\usepackage{graphicx}
\usepackage{epstopdf}
\usepackage{amsmath}
\usepackage{url}
\usepackage{amssymb}

\begin{document}
\title{SEBERTNets: Sequence Enhanced BERT Networks for Event Entity Extraction Tasks Oriented to the Finance Field}
%
%

\author{Congqing He\inst{1} \and
Xiangyu Zhu\inst{1} \and
Yuquan Le\inst{1}\and
Yuzhong Liu\inst{1} 
Jianhong Yin\inst{1} 
}

\authorrunning{He et al.}
%
\institute{JD Digital \\
\email{hecongqing@hotmail.com}}

%
\maketitle              
\begin{abstract}
Event extraction lies at the cores of investment analysis and asset management in the financial field, and thus has received much attention. The 2019 China conference on knowledge graph and semantic computing (CCKS) challenge sets up a evaluation competition for event entity extraction task oriented to the finance field. In this task, we mainly focus on how to extract the event entity accurately, and recall all the corresponding event entity effectively. In this paper, we propose a novel model, \textbf{S}equence \textbf{E}nhanced \textbf{BERT} \textbf{Net}works (SEBERTNets for short), which can inherit the advantages of the BERT,and while capturing sequence semantic information. In addition, motivated by recommendation system, we propose Hybrid Sequence Enhanced BERT Networks (HSEBERTNets for short), which uses a multi-channel recall method to recall all the corresponding event entity. The experimental results show that, the F1 score of SEBERTNets is 0.905 in the first stage, and the F1 score of HSEBERTNets is 0.934 in the first stage, which demonstarate the effectiveness of our methods.

\end{abstract}
\section{Introduction}
Event Extraction, a challenging task Information Extraction, aims at detecting and typing events, and extracting different entity from texts. Event extraction lies at the cores of investment analysis and asset management in the financial field, and thus has received much attention. However, most of research is rarely involved in the finance field. For this purpose, the 2019 China conference on knowledge graph and semantic computing (CCKS) challenge sets up a evaluation competition for event entity extraction task oriented to the finance field. The goal of the evaluation is to extract the event entity according to  a given real news corpus and event type. 

To this end, Liu et al.~\cite{liu2017exploiting}~ proposed  to exploit argument information explicitly for event detection via supervised attention mechanisms. There are still several questions. 
(1) How to extract the event entity  accurately and effectively. 
(2) How to recall all the corresponding event entity effectively when there exists multiple event entity in a text.
Motivated by these, we propose a novel model, Sequence Enhanced BERT Networks, which can inherit the advantages of the BERT~\cite{devlin-etal-2019-bert}, and while capturing sequence semantic information. In addition, motivated by recommendation system, we propose Hybrid Sequence Enhanced BERT Networks , which uses a multi-channel recall method to recall all the corresponding event entity.

Our mainly contributions are shown as follows:

(1) We propose a novel model, Sequence Enhanced BERT Networks, which can inherit the advantages of the BERT,and while capturing sequence semantic information.

(2) We propose Hybrid Sequence Enhanced BERT Networks , which uses a multi-channel recall method to recall all the corresponding event entity.

(3) The experimental results show that, the F1 score of SEBERTNets is 0.905 in the first stage, and the F1 score of HSEBERTNets is 0.934 in the first stage, which demonstarate the effectiveness of our methods.

\section{Our Methods}

\begin{figure}[ht]
\includegraphics[width=0.8\textwidth]{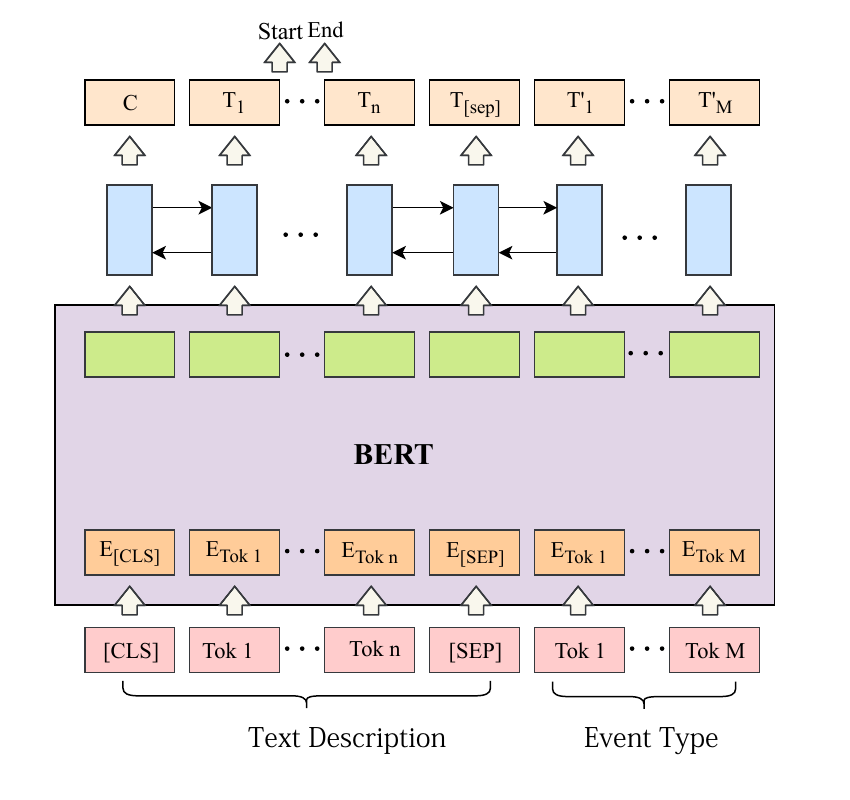}
\caption{The architecture of SEBERTNets.} \label{fig.model}
\end{figure}

Figure \ref{fig.model} describes the architecture of SEBERTNets, which primarily involves the following four components: (i) Input layer, (ii) BERT layer, (iii) Sequence layer, (iv) Output layer.

\subsection{Input Layer}

The description of text can be seen as a char sequence $\text{x}={\{x_1, \cdots, x_n\}}$, and the corresponding event type can be seen as a char sequence $\text{t}={\{t_1, \cdots, t_m\}}$. We first remove irregular punctuation, special text, etc, and  contact them together. The description of text $\text{x}$ and the corresponding event type $\text{t}$
are then encoded as input.

\subsection{BERT Layer}

Inspired by Devlin et al \cite{devlin-etal-2019-bert}, BERT obtains new state-of-the-art results on lots of natural language processing tasks. We introduce BERT as the layer of SEBERTNets, to produce sequence semantic representation, and capture global semantic representation.

\begin{equation}
{l_1, l_2, \cdots, l_{n+m}}= \textbf{BERT}(x_1, \cdots, x_n, t_1, \cdots, t_m)
\end{equation}

Where $l_i \in \mathbb{R}^{d}$,~\footnote{In our experiment, d=768.} and $i$ represents $i-th$ element.

\subsection{Sequence Layer}
Since the BERT layer only captures the position information of the text through the position vector, then the SEBERTNets introduces the Sequence layer, which captures the sequence semantic information of the text, and enhances the semantics of the text sequence based on the BERT layer. In addition, the Sequence layer introduces a mask operation. Since the text is usually indefinitely long, the introduction of a mask can effectively alleviate the bias caused by the text filling.

\begin{align}
\label{eq:lstm}
&f_t=\sigma(W_{f}l_{t}+U_{f}h_{t-1}+b_{f}),\\
&i_t=\sigma(W_{i}l_{t}+U_{i}h_{t-1}+b_{i}),\\
&o_t=\sigma(W_{o}l_{t}+U_{o}h_{t-1}+b_{o}),\\
&\widetilde{c}_{t}=tanh(W_{c}x_{t}+U_{c}h_{t-1}+b_{c}),\\
&c_{t}=f_{t}\odot c_{t-1}+i_t\odot \widetilde{c}_{t},\\
&h_{t}=o_{t}\odot tanh(c_{t}).
\end{align}

Where $l_t$ represents the input of the recurrent layer at time step $t$. $f_t$, $i_t$ and $o_t$ means forget gate, input gate, and output gate respectively. $\odot$ denotes element-wise multiplication of two vectors. To consider forward and backward contextual representation of the text, we use BiLSTM  instead of LSTM as fact encoder in advance. The BiLSTM generates ${h_t}_{\{t=1, \cdots, T\}}$ by concatenating a forward LSTM and a backward LSTM.

\subsection{Output Layer}
The BERT layer and the Sequence layer can capture the global semantic information and sequence semantic information of the text. Based on the text and event type, the SEBERTNets model predicts the beginning and end of the text of the event body. Finally, it is decoded by the start position and the end position to obtain the corresponding event body.

\subsection{Optimization}

Although Adam~\cite{kingma2014adam}~ performed well at the beginning of model training, SGDStochastic gradient descent (SGD)~\cite{robbins1951stochastic}~ was superior to the Adam method in the later stages of training. Therefore, we introduce a new method, SWATS~\cite{keskar2017improving}~, which first starts training with the Adam method and switches to SGD when appropriate. Verification on the real data set of the evaluation game, this method can achieve better performance than the Adam optimization method.

\section{Experiments}

\subsection{Dataset}
CCKS 2019 evaluation task  consists of 17,815 training sets, 3,500 validation sets (preliminary test datasets) and 135,519 test sets (rematch test datasets). Each dataset includes a given text description, event type, and event entity respectively.

\subsection{Evaluation}

Following~\cite{hong2011using}~, we use F1 score to evaluate the performance of our methods. The F1 score is computed as follows:

\begin{equation}
F1 = \frac{2PR}{P+R}
\end{equation}
Where $P$ means that number of correctly identified event entities divided by the total number of identified event entities, $R$ represents that number of correctly identified event entities divided by the total number of annotated event entities.

\subsection{Hyper-parameter Setting}

Since all the text descriptions and event types have been employed  for char segmentation and set each text maximum length to 140. We use character-based Chinese BERT model~\footnote{\url{https://storage.googleapis.com/bert_models/2018_11_03/chinese_L-12_H-768_A-12.zip}}~and character-based Chinese BERT-wwm model~\footnote{\url{https://drive.google.com/open?id=1RoTQsXp2hkQ1gSRVylRIJfQxJUgkfJMW}}~to training SEBERTNets and variant SEBERTNets models. In addition, GRU is adopted to Sequence layer and the hidden unit is set to 200. The batch size is set to 32 for SEBERTNets and variant SEBERTNets models.

\section{Results}

\subsection{Main Results}

\begin{table}[htbp]
\centering
\caption{F1 score of models on preliminary test datasets.}
\label{tab.mainresult}
\setlength{\tabcolsep}{3mm}{
\begin{tabular}{cccccc}
\hline
Number of entities& Top 1 &Top 2  &Top 3 &Top 4 &Top 5 \cr
\hline
BERT&	0.871&	0.887&	0.893&	0.901&	0.902\\
SEBERTNets&	0.905&	0.915&	0.918&	0.921&	0.923\\
HSEBERTNets&	\bf{0.914}&	\bf{0.922}&	\bf{0.926}&	\bf{0.931}&	\bf{0.934}\\
\hline
\end{tabular}
}
\end{table}
In this section, we conduct experiments on preliminary test datasets to demonstrate the effectiveness of the proposed approach. Table~\ref{tab.mainresult}~ shows the performance of our methods on preliminary test datasets.

Overall, we can find the SEBERTNets model outperforms BERT model with a significant margin on preliminary test datasets. SEBERTNets model obtains $2.5\%$ average absolutely considerable improvements on preliminary test datasets, which demonstrates the effectiveness of SEBERTNets model for event entity extraction.

Additionally, we propose HSEBERTNets model to recall all the corresponding event entity. Compared to the SEBERTNets model, HSEBERTNets model achieves better performance, especially, in multiple event entities. The results demonstrates the effectiveness of HSEBERTNets model for multiple event entity extraction.

\subsection{Visualization}

\begin{figure}[ht]
\label{fig.vvv}
\includegraphics[width=\textwidth]{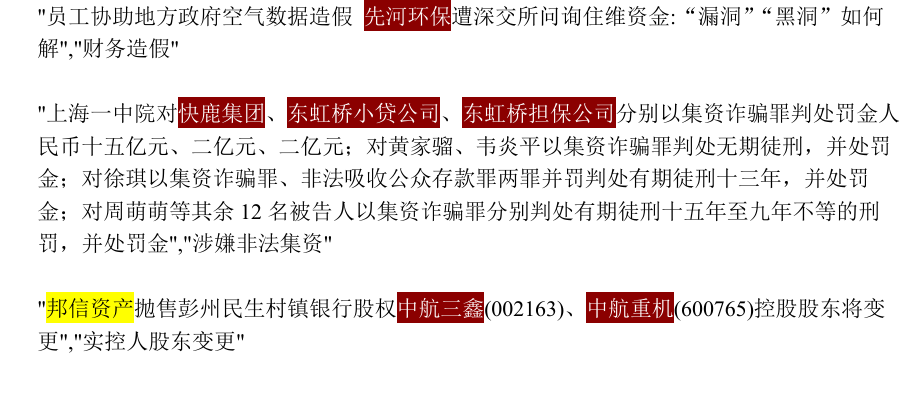}
\caption{Visualization of BERT layer.} 
\label{fig1}
\end{figure}

In this section, we select several representative cases to give an intuitive illustration of how the BERT layer help to promote the performance of event entity extraction. As shown in Figure~\ref{fig1}, red color means that the correct entities extracted by SEBERTNets model, and yellow color means that the incorrect entites extracted by SEBERTNets model. In these cases, we can find that BERT layer focus on the entity of the text well. Several entities are not belonging to the corresponding event type, and BERT layer can be well observed. From this figure, we observe that the BERT layer can capture key entities  relevant to current event type.

%
%
\bibliographystyle{splncs04}
\bibliography{mybibliography}

\begin{thebibliography}{1}
\providecommand{\url}[1]{\texttt{#1}}
\providecommand{\urlprefix}{URL }
\providecommand{\doi}[1]{https://doi.org/#1}

\bibitem{devlin-etal-2019-bert}
Devlin, J., Chang, M.W., Lee, K., Toutanova, K.: {BERT}: Pre-training of deep
  bidirectional transformers for language understanding. In: Proceedings of the
  2019 Conference of the North {A}merican Chapter of the Association for
  Computational Linguistics: Human Language Technologies, Volume 1 (Long and
  Short Papers). pp. 4171--4186. Association for Computational Linguistics,
  Minneapolis, Minnesota (Jun 2019). \doi{10.18653/v1/N19-1423},
  \url{https://www.aclweb.org/anthology/N19-1423}

\bibitem{hong2011using}
Hong, Y., Zhang, J., Ma, B., Yao, J., Zhou, G., Zhu, Q.: Using cross-entity
  inference to improve event extraction. In: Proceedings of the 49th Annual
  Meeting of the Association for Computational Linguistics: Human Language
  Technologies-Volume 1. pp. 1127--1136. Association for Computational
  Linguistics (2011)

\bibitem{keskar2017improving}
Keskar, N.S., Socher, R.: Improving generalization performance by switching
  from adam to sgd. arXiv preprint arXiv:1712.07628  (2017)

\bibitem{kingma2014adam}
Kingma, D.P., Ba, J.: Adam: A method for stochastic optimization. arXiv
  preprint arXiv:1412.6980  (2014)

\bibitem{liu2017exploiting}
Liu, S., Chen, Y., Liu, K., Zhao, J., et~al.: Exploiting argument information
  to improve event detection via supervised attention mechanisms  (2017)

\bibitem{robbins1951stochastic}
Robbins, H., Monro, S.: A stochastic approximation method. The annals of
  mathematical statistics pp. 400--407 (1951)

\end{thebibliography}

\end{document}